\documentclass[11pt,a4paper]{article}
\usepackage[hyperref]{emnlp-ijcnlp-2019}
\usepackage{times}
\usepackage{enumitem}
\usepackage{latexsym}
\usepackage{graphicx}
\usepackage{url}

\aclfinalcopy % Uncomment this line for the final submission
\setlist{noitemsep,topsep=10pt,parsep=0pt,partopsep=0pt}

\newcommand*{\subscript}[1]{\ensuremath{_\textrm{{\scriptsize #1}}}}

\title{DENS: A Dataset for Multi-class Emotion Analysis}

\author{Chen Liu \and
  Muhammad Osama \and
  Anderson de Andrade \\
  Wattpad\\
  Toronto, ON, Canada \\
  \texttt{cecilia, muhammad.osama, anderson@wattpad.com}
}

\date{}

\begin{document}
\maketitle
\begin{abstract}
	We introduce a new dataset for multi-class emotion analysis from long-form narratives in English. The Dataset for Emotions of Narrative Sequences (DENS) was collected from both classic literature available on Project Gutenberg and modern online narratives available on Wattpad, annotated using Amazon Mechanical Turk. A number of statistics and baseline benchmarks are provided for the dataset. Of the tested techniques, we find that the fine-tuning of a pre-trained BERT model achieves the best results, with an average micro-F1 score of 60.4\%. Our results show that the dataset provides a novel opportunity in emotion analysis that requires moving beyond existing sentence-level techniques. 
\end{abstract}

\section{Introduction}
Humans experience a variety of complex emotions in daily life. These emotions are heavily reflected in our language, in both spoken and written forms. 

Many recent advances in natural language processing on emotions have focused on product reviews ~\cite{AmazonProductDataset} and tweets ~\cite{SemEval1,MultiemotionVideoGameTweetDataset}. These datasets are often limited in length (e.g. by the number of words in tweets), purpose (e.g. product reviews), or emotional spectrum (e.g. binary classification).

Character dialogues and narratives in storytelling usually carry strong emotions. A memorable story is often one in which the emotional journey of the characters resonates with the reader. Indeed, emotion is one of the most important aspects of narratives. In order to characterize narrative emotions properly, we must move beyond binary constraints (e.g. good or bad, happy or sad).

In this paper, we introduce the Dataset for Emotions of Narrative Sequences (DENS) for emotion analysis, consisting of passages from long-form fictional narratives from both classic literature and modern stories in English. The data samples consist of self-contained passages that span several sentences and a variety of subjects. Each sample is annotated by using one of 9 classes and an indicator for annotator agreement. 

\section{Background}

Using the categorical basic emotion model~\cite{Plutchik:79}, ~\cite{EmoTweet1, EmoTweet2} studied creating lexicons from tweets for use in emotion analysis. Recently, ~\cite{SemEval1}, ~\cite{WASSA2018} and ~\cite{MultiemotionVideoGameTweetDataset} proposed shared-tasks for multi-class emotion analysis based on tweets.

Fewer works have been reported on understanding emotions in narratives. Emotional Arc ~\cite{Reagan2016} is one recent advance in this direction. The work used lexicons and unsupervised learning methods based on unlabelled passages from titles in Project Gutenberg\footnote{\url{https://www.gutenberg.org/}}.

For labelled datasets on narratives, \cite{Alm:05} provided a sentence-level annotated corpus of childrens' stories and ~\cite{C18-1114} provided phrase-level annotations on selected Project Gutenberg titles.

To the best of our knowledge, the dataset in this work is the first to provide multi-class emotion labels on passages, selected from both Project Gutenberg and modern narratives. The dataset is available upon request for non-commercial, research only purposes\footnote{Please send requests to: academic\_dataset@wattpad.com}.

\section{Dataset}
In this section, we describe the process used to collect and annotate the dataset. 

\subsection{Plutchik’s Wheel of Emotions}
The dataset is annotated based on a modified Plutchik’s wheel of emotions.

The original Plutchik’s wheel consists of 8 primary emotions: \textit{Joy, Sadness, Anger, Fear, Anticipation, Surprise, Trust, Disgust}. In addition, more complex emotions can be formed by combing two basic emotions. For example, \textit{Love} is defined as a combination of \textit{Joy} and \textit{Trust} (Fig. 1).

\begin{figure}[h]
\includegraphics[width=0.5\textwidth]{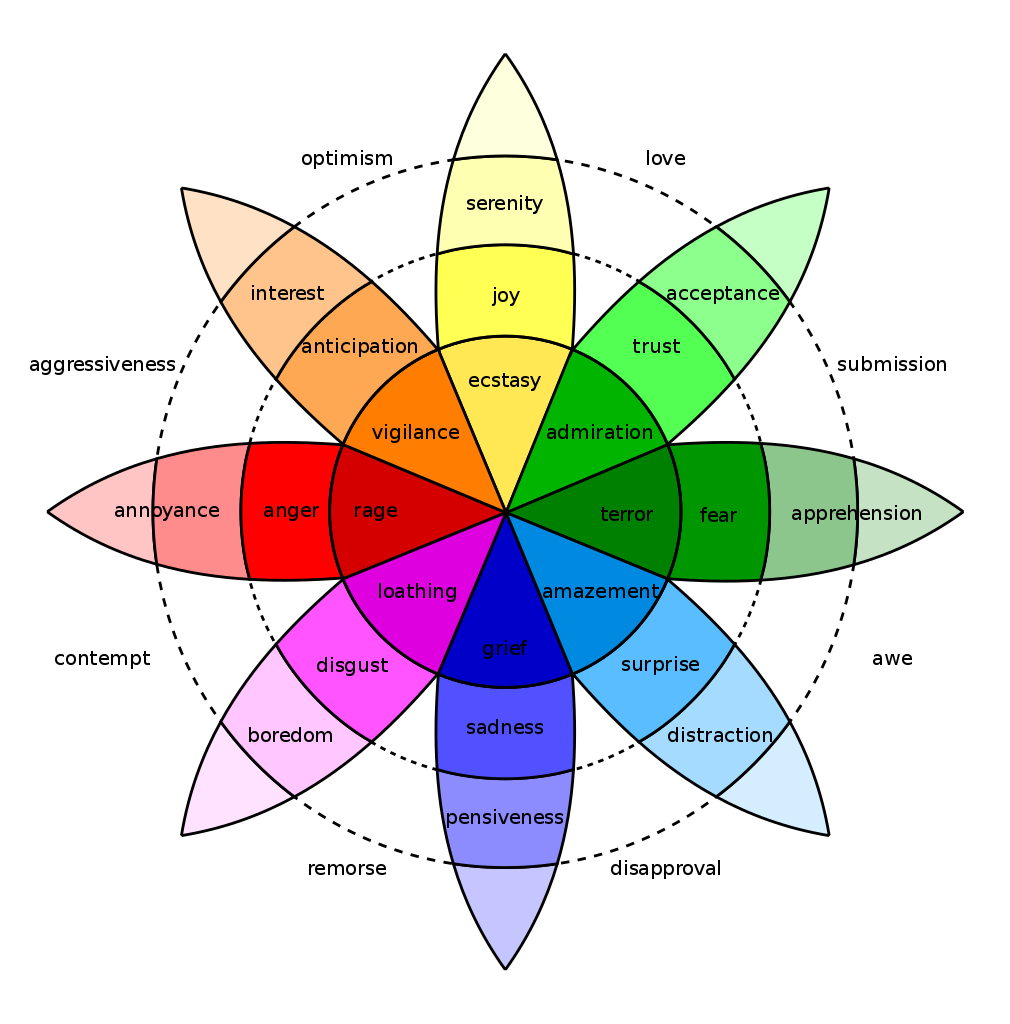}
\caption{Plutchik's wheel of emotions ~\cite{PlutchikChart}}
\end{figure}

The intensity of an emotion is also captured in Plutchik's wheel. For example, the primary emotion of \textit{Anger} can vary between \textit{Annoyance} (mild) and \textit{Rage} (intense).

We conducted an initial survey based on 100 stories with a significant fraction sampled from the romance genre. We asked readers to identify the major emotion exhibited in each story from a choice of the original 8 primary emotions. 

We found that readers have significant difficulty in identifying \textit{Trust} as an emotion associated with romantic stories. Hence, we modified our annotation scheme by removing \textit{Trust} and adding \textit{Love}. We also added the \textit{Neutral} category to denote passages that do not exhibit any emotional content. 

The final annotation categories for the dataset are:  \textit{Joy, Sadness, Anger, Fear, Anticipation, Surprise, Love, Disgust, Neutral}. 

\subsection{Passage Selection}
We selected both classic and modern narratives in English for this dataset.  The modern narratives were sampled based on popularity from Wattpad. We parsed selected narratives into passages, where a passage is considered to be eligible for annotation if it contained between 40 and 200 tokens. 

In long-form narratives, many non-conversational passages are intended for transition or scene introduction, and may not carry any emotion. We divided the eligible passages into two parts, and one part was pruned using selected emotion-rich but ambiguous lexicons such as \textit{cry, punch, kiss, etc.}. Then we mixed this pruned part with the unpruned part for annotation in order to reduce the number of neutral passages. See Appendix \ref{sec:lexicon} for the lexicons used.

\subsection{Mechanical Turk (MTurk)}

MTurk was set up using the standard sentiment template and instructed the crowd annotators to `pick the best/major emotion embodied in the passage'.

We further provided instructions to clarify the intensity of an emotion, such as: ``\textit{Rage}/\textit{Annoyance} is a form of \textit{Anger}'', ``\textit{Serenity}/\textit{Ecstasy} is a form of \textit{Joy}'', and ``\textit{Love} includes \textit{Romantic}/\textit{Family}/\textit{Friendship}'', along with sample passages.

We required all annotators have a `master' MTurk qualification. Each passage was labelled by 3 unique annotators. Only passages with a majority agreement between annotators were accepted as valid. This is equivalent to a Fleiss's $\kappa$ score of greater than $0.4$.

For passages without majority agreement between annotators, we consolidated their labels using in-house data annotators who are experts in narrative content. A passage is accepted as valid if the in-house annotator's label matched any one of the MTurk annotators' labels. The remaining passages are discarded. We provide the fraction of annotator agreement for each label in the dataset. 

Though passages may lose some emotional context when read independently of the complete narrative, we believe annotator agreement on our dataset supports the assertion that small excerpts can still convey coherent emotions.

During the annotation process, several annotators had suggested for us to include additional emotions such as confused, pain, and jealousy, which are common to narratives. As they were not part of the original Plutchik’s wheel, we decided to not include them. An interesting future direction is to study the relationship between emotions such as ‘pain versus sadness’ or ‘confused versus surprise’ and improve the emotion model for narratives.

\subsection{Dataset Statistics}
The dataset contains a total of 9710 passages, with an average of 6.24 sentences per passage, 16.16 words per sentence, and an average length of 86 words. 

The vocabulary size is 28K (when lowercased). It contains over 1600 unique titles across multiple categories, including 88 titles (1520 passages) from Project Gutenberg. All of the modern narratives were written after the year 2000, with notable amount of themes in \textit{coming-of-age}, \textit{strong-female-lead}, and \textit{LGBTQ+}. The genre distribution is listed in Table~\ref{genre_stats}. 

\begin{table}
	\centering
	\begin{tabular}{lr}
		Genre  & Distribution (\%)\\
		\hline
		Mystery/Thriller & 19.7 \\
		Paranormal & 16.6 \\
		Fantasy & 13.2 \\
		Horror & 11.3 \\
		Romance & 8.7 \\
		Action/Adventure & 5.3 \\
		\hline
		Other & 9.3 \\
	\end{tabular}
	\caption{\label{genre_stats} Genre distribution of the modern narratives}
\end{table} 

In the final dataset, 21.0\% of the data has consensus between all annotators, 73.5\% has majority agreement, and 5.48\% has labels assigned after consultation with in-house annotators.

The distribution of data points over labels with top lexicons (lower-cased, normalized) is shown in Table~\ref{label_stats}. Note that the \textit{Disgust} category is very small and should be discarded. Furthermore, we suspect that the data labelled as \textit{Surprise} may be noisier than other categories and should be discarded as well.

\begin{table*}
	\centering
	\begin{tabular}{lrrl}
		Label & Gutenberg & Total & Top Lexicons\\
		\hline
		Neutral & 318 & 1711 & take, love, long, really, want, always, though, away, look\\
		Fear & 159 & 1412 & left, behind, right, want, let, death, go, say, think\\
		Sadness & 195 & 1402 & father, always, little, look, something, us, really, mother, think\\
		Anger & 192 & 1306  & feel, much, well, man, look, us, say, something, love \\
		Joy & 241 & 1266 & see, always, let, long, make, hand, away, get, really\\
		Love & 162 & 1157 & hand, know, right, let, happy, get, ever, us, look \\
		Anticipation & 147 & 1020 & know, long, life, make, get, think, blood, want, feel \\
		Surprise & 102 & 362 & love, find, looking, know, well, much, something, door, really\\
		Disgust & 4 & 74 & get, hand, inside, let, hate, table, men, always, make\\
	\end{tabular}
	\caption{\label{label_stats} Dataset label distribution}
\end{table*} 

Table~\ref{sample_data_small} shows a few examples labelled data from classic titles. More examples can be found in Table~\ref{sample_data} in the Appendix \ref{sec:sample_data}.

\begin{table*}[h!]
	\centering
	\begin{tabular}{p{12cm}|p{2cm}}
	Text & Label \\
		\hline
		\textit{I found this was a little too close upon him, but I made it up in what follows. He stood stock-still for a while and said nothing, and I went on thus: ``You cannot," says I, `without the highest injustice, believe that I yielded upon all these persuasions without a love not to be questioned, not to be shaken again by anything that could happen afterward. If you have such dishonourable thoughts of me, I must ask you what foundation in any of my behaviour have I given for such a suggestion?"} & Angry  \\\hline
		\textit{She stretched hers eagerly and gratefully towards him. What had happened? Through all the numbness of her blood, there sprang a strange new warmth from his strong palm, and a pulse, which she had almost forgotten as a dream of the past, began to beat through her frame. She turned around all a-tremble, and saw his face in the glow of the coming day.} & Anticipation\\\hline
		\textit{Ah! That moving procession that has left me by the road-side! Its fantastic colors are more brilliant and beautiful than the sun on the undulating waters. What matter if souls and bodies are failing beneath the feet of the ever-pressing multitude! It moves with the majestic rhythm of the spheres. Its discordant clashes sweep upward in one harmonious tone that blends with the music of other worlds--to complete God's orchestra.} & Joy  \\\hline
	\end{tabular}
	\caption{\label{sample_data_small} Sample data from classic titles}
\end{table*}

\section{Benchmarks}
We performed benchmark experiments on the dataset using several different algorithms. In all experiments, we have discarded the data labelled with \textit{Surprise} and \textit{Disgust}. 

We pre-processed the data by using the SpaCy\footnote{\url{https://spacy.io/}} pipeline. We masked out named entities with entity-type specific placeholders to reduce the chance of benchmark models utilizing named entities as a basis for classification. 

Benchmark results are shown in Table~\ref{benchmark}. The dataset is approximately balanced after discarding the \textit{Surprise} and \textit{Disgust} classes. We report the average micro-F1 scores, with 5-fold cross validation for each technique. 

We provide a brief overview of each benchmark experiment below. Among all of the benchmarks, Bidirectional Encoder Representations from Transformers (BERT) ~\cite{BERT} achieved the best performance with a 0.604 micro-F1 score.

Overall, we observed that deep-learning based techniques performed better than lexical based methods. This suggests that a method which attends to context and themes could do well on the dataset.

\subsection{Bag-of-Words-based Benchmarks}
We computed bag-of-words-based benchmarks using the following methods:

\begin{itemize}
	\item Classification with TF-IDF + Linear SVM (TF-IDF + SVM)
	\item Classification with Depeche++ Emotion lexicons ~\cite{Depeche} + Linear SVM (Depeche + SVM)
	\item Classification with NRC Emotion lexicons ~\cite{Lexicon2, Lexicon1} + Linear SVM (NRC + SVM)
	\item Combination of TF-IDF and NRC Emotion lexicons (TF-NRC + SVM)
\end{itemize}

\begin{table}[t!]
	\centering
	\begin{tabular}{lr}
		Model & micro-F1\\
		\hline
		TF-IDF + SVM & 0.450 \\
		Depeche + SVM & 0.254 \\
		NRC + SVM & 0.286 \\
		TF-NRC + SVM & 0.458  \\\hline
		Doc2Vec + SVM & 0.403 \\
		HRNN & 0.469 \\
		BiRNN + Self-Attention & 0.487 \\
		ELMo + BiRNN & 0.516\\
		Fine-tuned BERT & \textbf{0.604}
	\end{tabular}
	\caption{\label{benchmark}Benchmark results (averaged 5-fold cross validation)}
\end{table} 

\subsection{Doc2Vec + SVM}

We also used simple classification models with learned embeddings. We trained a Doc2Vec model~\cite{D2V} using the dataset and used the embedding document vectors as features for a linear SVM classifier.

\subsection{Hierarchical RNN}

For this benchmark, we considered a Hierarchical RNN, following~\cite{Hierarchical}. We used two BiLSTMs ~\cite{BiLSTM} with 256 units each to model sentences and documents. The tokens of a sentence were processed independently of other sentence tokens. For each direction in the token-level BiLSTM, the last outputs were concatenated and fed into the sentence-level BiLSTM as inputs. 

The outputs of the BiLSTM were connected to 2 dense layers with 256 ReLU units and a Softmax layer. We initialized tokens with publicly available embeddings trained with GloVe ~\cite{Glove}. Sentence boundaries were provided by SpaCy. Dropout was applied to the dense hidden layers during training.
  
\subsection{Bi-directional RNN and Self-Attention (BiRNN + Self-Attention)}

One challenge with RNN-based solutions for text classification is finding the best way to combine word-level representations into higher-level representations. 

Self-attention~\cite{ATT:16, SSA:17, ATT:18} has been adapted to text classification, providing improved interpretability and performance. We used~\cite{SSA:17} as the basis of this benchmark. 

The benchmark used a layered Bi-directional RNN (60 units) with GRU cells and a dense layer. Both self-attention layers were 60 units in size and cross-entropy was used as the cost function. 

Note that we have omitted the orthogonal regularizer term, since this dataset is relatively small compared to the traditional datasets used for training such a model. We did not observe any significant performance gain while using the regularizer term in our experiments.

\subsection{ELMo embedding and Bi-directional RNN (ELMo + BiRNN)}

Deep Contextualized Word Representations (ELMo)~\cite{ELMo} have shown recent success in a number of NLP tasks. The unsupervised nature of the language model allows it to utilize a large amount of available unlabelled data in order to learn better representations of words.

We used the pre-trained ELMo model (v2) available on \textit{Tensorhub}\footnote{\url{https://tfhub.dev/google/elmo/2}} for this benchmark. We fed the word embeddings of ELMo as input into a one layer Bi-directional RNN (16 units) with GRU cells (with dropout) and a dense layer. Cross-entropy was used as the cost function.

\subsection{Fine-tuned BERT}

Bidirectional Encoder Representations from Transformers (BERT) ~\cite{BERT} has achieved state-of-the-art results on several NLP tasks, including sentence classification.

We used the fine-tuning procedure outlined in the original work to adapt the pre-trained uncased
 BERT\subscript{LARGE}\footnote{\url{https://tfhub.dev/google/bert_uncased_L-24_H-1024_A-16/1}} to a multi-class passage classification task. This technique achieved the best result among our benchmarks, with an average micro-F1 score of 60.4\%.

\section{Conclusion}

We introduce DENS, a dataset for multi-class emotion analysis from long-form narratives in English. We provide a number of benchmark results based on models ranging from bag-of-word models to methods based on pre-trained language models (ELMo and BERT). 

Our benchmark results demonstrate that this dataset provides a novel challenge in emotion analysis. The results also demonstrate that attention-based models could significantly improve performance on classification tasks such as emotion analysis. 

Interesting future directions for this work include: 1. incorporating common-sense knowledge into emotion analysis to capture semantic context and 2. using few-shot learning to bootstrap and improve performance of underrepresented emotions.

Finally, as narrative passages often involve interactions between multiple emotions, one avenue for future datasets could be to focus on the multi-emotion complexities of human language and their contextual interactions.

% Min: no longer used as of ACL 2018, following ACL exec's decision to
% remove this extra workflow that was not executed much.
% BEGIN: remove
%% \section{XML conversion and supported \LaTeX\ packages}

%% Following ACL 2014 we will also we will attempt to automatically convert 
%% your \LaTeX\ source files to publish papers in machine-readable 
%% XML with semantic markup in the ACL Anthology, in addition to the 
%% traditional PDF format.  This will allow us to create, over the next 
%% few years, a growing corpus of scientific text for our own future research, 
%% and picks up on recent initiatives on converting ACL papers from earlier 
%% years to XML. 

%% We encourage you to submit a ZIP file of your \LaTeX\ sources along
%% with the camera-ready version of your paper. We will then convert them
%% to XML automatically, using the LaTeXML tool
%% (\url{http://dlmf.nist.gov/LaTeXML}). LaTeXML has \emph{bindings} for
%% a number of \LaTeX\ packages, including the ACL 2018 stylefile. These
%% bindings allow LaTeXML to render the commands from these packages
%% correctly in XML. For best results, we encourage you to use the
%% packages that are officially supported by LaTeXML, listed at
%% \url{http://dlmf.nist.gov/LaTeXML/manual/included.bindings}
% END: remove

%%\section*{Acknowledgments}

%%The acknowledgments should go immediately before the references.  Do
%%not number the acknowledgments section. Do not include this section
%%when submitting your paper for review. \\

%%\noindent \textbf{Preparing References:} \\
%%Include your own bib file like this:
%%\verb|\bibliographystyle{acl_natbib}|
%%\verb|\bibliography{acl2019}| 

%%where \verb|acl2019| corresponds to a acl2019.bib file.

\bibliography{acl2019}
\bibliographystyle{acl_natbib}

%Appendices should be \textbf{uploaded as supplementary material} when submitting the paper for review. Upon acceptance, the appendices come after the references, as shown here. Use \verb|\appendix| before any appendix section to switch the section numbering over to letters.

\appendix
\onecolumn
\section{Appendices}
%\label{sec:appendix}
\subsection{Lexicons}
\label{sec:lexicon}

\begin{table*}[h!]
	\centering
	\begin{tabular}{llll}
		cry & punch & blood & knife \\
		flower & moon &	wind & exclaim \\
		chuckle & tear & punch & yell \\
		kiss & touch & warm & dead \\
		shiver & chill \\
	\end{tabular}
\caption{\label{sample_data} Lexicons used to prune part of the data for labelling}
\end{table*}

\subsection{Sample Data}
\label{sec:sample_data}
Table \ref{sample_data} shows sample passages from classic titles with corresponding labels. 

\begin{table*}[h!]
	\centering
	\begin{tabular}{p{12cm}|p{1cm}}
		Text & Label \\
		\hline
		\textit{He took his screwdriver and again took off the lid of the coffin. Arthur looked on, very pale but silent. When the lid was removed he stepped forward. He evidently did not know that there was a leaden coffin, or at any rate, had not thought of it. When he saw the rent in the lead, the blood rushed to his face for an instant, but as quickly fell away again, so that he remained of a ghastly whiteness. He was still silent. Van Helsing forced back the leaden flange, and we all looked in and recoiled.} & Fear \\\hline
		\textit{The chair went to matchwood at the bottom, and we rolled apart into the gutter. He sprang to his feet, waving his fists and wheezing like an asthmatic. ``Had enough?" he panted. ``You infernal bully!" I cried, as I gathered myself together.} & Anger \\\hline
		\textit{The judges sat grave and mute, gave me an easy hearing, and time to say all that I would, but, saying neither Yes nor No to it, pronounced the sentence of death upon me, a sentence that was to me like death itself, which, after it was read, confounded me. I had no more spirit left in me, I had no tongue to speak, or eyes to look up either to God or man.} & Sadness \\\hline
		\textit{The Prince burst into a yelling, shrieking fit of laughter. Instantly the yellow-haired serfs in waiting, the Calmucks at the hall-door, and the half-witted dwarf who crawled around the table in his tow shirt, began laughing in chorus, as violently as they could. The Princess Martha and Prince Boris laughed also; and while the old man's eyes were dimmed with streaming tears of mirth, quickly exchanged nods. The sound extended all over the castle, and was heard outside of the walls.} & Joy \\\hline
		\textit{``Do not be such an unreasonable child", he remonstrated, feebly. ``I do not love you with the wild, irrational passion of former years; but I have the tenderest regard for you, and my heart warms at the sight of your sweet face, and I shall do all in my power to make you as happy as any man can make you who--"} & Love \\\hline
		\textit{I looked around for his birds, and not seeing them, asked him where they were. He replied, without turning round, that they had all flown away. There were a few feathers about the room and on his pillow a drop of blood. I said nothing, but went and told the keeper to report to me if there were anything odd about him during the day.} & Neutral \\
	\end{tabular}
	\caption{\label{sample_data} Sample data from classic titles}
\end{table*}
\end{document}